\documentclass[10pt,twocolumn,letterpaper]{article}

\usepackage[pagenumbers]{cvpr} %

\usepackage{amsmath,amsfonts,bm}

\def\eqref#1{equation~\ref{#1}}

\def\1{\bm{1}}

\DeclareMathAlphabet{\mathsfit}{\encodingdefault}{\sfdefault}{m}{sl}
\SetMathAlphabet{\mathsfit}{bold}{\encodingdefault}{\sfdefault}{bx}{n}

\usepackage{graphicx}
\usepackage{amssymb}
\usepackage{booktabs}

\usepackage{url}            %
\usepackage{nicefrac}       %
\usepackage{microtype}      %
\usepackage{xcolor}         %
\usepackage{makecell}
\usepackage{multirow}
\usepackage{color, colortbl}
\usepackage{enumitem}
\usepackage{subcaption}
\usepackage{caption}
\usepackage[utf8]{inputenc} %
\usepackage[T1]{fontenc}    %
\usepackage{threeparttable} %
\usepackage{enumitem}
\usepackage{mathtools}

\newcommand{\z}{\mathcal{Z}}

\usepackage[pagebackref,breaklinks,colorlinks]{hyperref}

\usepackage[capitalize]{cleveref}
\crefname{section}{Sec.}{Secs.}
\Crefname{section}{Section}{Sections}
\Crefname{table}{Table}{Tables}
\crefname{table}{Tab.}{Tabs.}

\newcommand{\method}[1]{\textit{GALS}}

\def\cvprPaperID{11622} %
\def\confName{CVPR}
\def\confYear{2022}

\begin{document}

\title{On Guiding Visual Attention with Language Specification}\vspace{-.4em}

 \author{Suzanne Petryk$^*$ \qquad
Lisa Dunlap$^*$ \qquad \\
Keyan Nasseri \qquad
Joseph Gonzalez \qquad 
Trevor Darrell \qquad
Anna Rohrbach\\ \\
\vspace{-.5em}
UC Berkeley
\vspace{-.8em}
}
\maketitle
\def\thefootnote{*}\footnotetext{Equal contribution.}\def\thefootnote{\arabic{footnote}}
\begin{abstract}

While real world challenges typically define visual categories with language words or phrases, most visual classification methods define categories with numerical indices. However, the language specification of the classes provides an especially useful prior for biased and noisy datasets, where it can help disambiguate what features are task-relevant. Recently, large-scale multimodal models have been shown to recognize a wide variety of high-level concepts from a language specification even without additional image training data, but they are often unable to distinguish classes for more fine-grained tasks. CNNs, in contrast, can extract subtle image features that are required for fine-grained discrimination, but will overfit to any bias or noise in datasets. Our insight is to use high-level language specification as {\em advice} for constraining the classification evidence to task-relevant features, instead of distractors. To do this, we ground task-relevant words or phrases with attention maps from a pretrained large-scale model. We then use this grounding to supervise a classifier's spatial attention away from distracting context. We show that supervising spatial attention in this way improves performance on classification tasks with biased and noisy data, including $\sim$3$-$15\% worst-group accuracy improvements and $\sim$41$-$45\% relative improvements on fairness metrics.

\end{abstract}

\section{Introduction}
\label{sec:intro}

\begin{figure}[t]
  \centering
  \includegraphics[width=\linewidth]{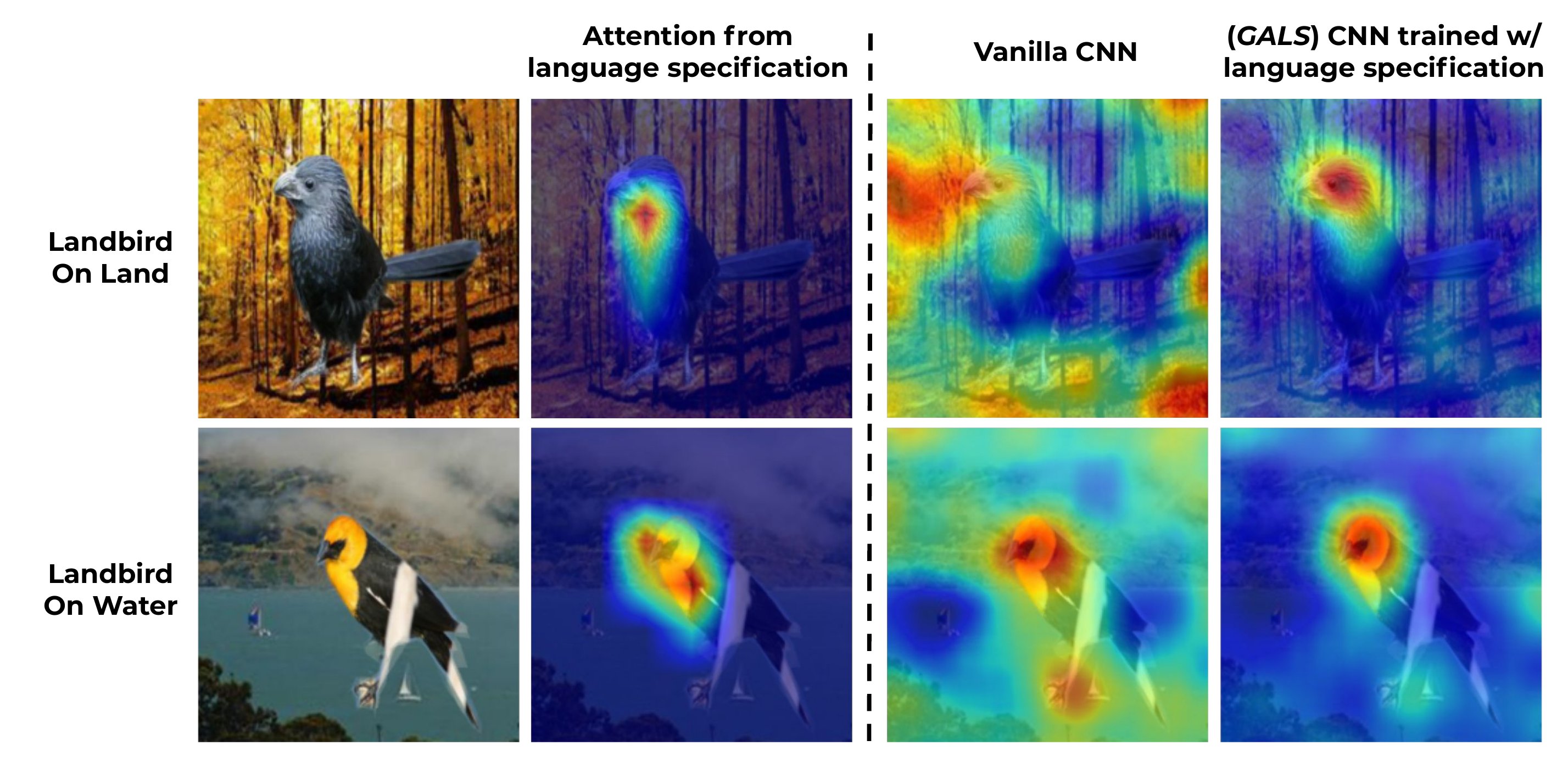}
  \caption{\textbf{Guiding attention with language.} Sample attention from the Waterbirds biased dataset. 
  During training, Landbirds mostly appear on land backgrounds and Waterbirds mostly appear on water backgrounds. At testing, each class appears equally on land or on water. A CNN trained on this task learns to look at the background, but if we use a multimodal model to translate the language specification \emph{``a photo of a bird''} into spatial supervision, we can ensure that our CNN learns task-relevant features.}
  \label{fig:teaser}
   \vspace{-2mm}
\end{figure}

When trained with limited or biased data, visual models often learn unwanted correlations.
For example, consider building a classifier to distinguish two fine-grained categories of birds: ``landbird'' and ``waterbird''. The background features from their corresponding habitats such as forests or beaches might be highly or perfectly correlated with the numerical class labels. 
A baseline model may mistakenly learn the unintended ``location'' task instead of the actual task, and fail on examples of birds out of their usual habitat (\cref{fig:teaser}). However, knowledge that the task is \emph{about birds} can disambiguate what the model is meant to learn.

Previous work has considered incorporating knowledge of the task as \emph{language specifications} in the form of class names or class descriptions which can directly serve as a prior over visual model parameters~\cite{radford2021learning,Wang_2019_CVPR}. Several zero-shot methods condition models on attribute labels \cite{lampert_zs, mall2021field, xu2020attribute} (e.g., beak shape or wing color)  or class descriptions \cite{lei2015predicting, elhoseiny2013write,qiao2016less,zhu2018generative} (e.g., from Wikipedia) to enable transfer to unseen classes. However, this relies on the language specification being class-discriminative -- an assumption which does not hold for some real-world tasks where only high-level task specification is given (e.g., in \cref{fig:teaser}, we may only know that this is a ``bird'' dataset, without the class names being provided or even existing yet). Additionally, simply conditioning on language embeddings may not prevent a model from attending to spurious correlations in biased datasets.

Even when language specifications \emph{are} class-discriminative, such models will perform poorly when there is insufficient image and text data to learn a multimodal model for rare or fine-grained classes (e.g., a large-scale model such as CLIP~\cite{radford2021learning} may not have seen enough examples of the relatively rare ``landbird'' or ``waterbird'' classes during pretraining to have good zero-shot performance).

To address these limitations, we propose a new framework called \textbf{Guiding visual Attention with Language Specification}, or \textbf{\method{}}, in which we translate available language specification provided by the task metadata into spatial attention that is used to supervise a CNN's attention during training.
\cref{fig:teaser} displays how \method{} is able to pull the model's attention away from the distractor features while retaining enough flexibility to pick up on fine-grained features which were not captured by the multimodal model.

Specifically, we first leverage an off-the-shelf pretrained vision-language model to \emph{ground} textual information into each given image and obtain a respective saliency map. This is efficient and involves no additional overhead (i.e., no need for training or per-instance manual annotation). Next, we aim to leverage the obtained saliency map to inform the visual classifier. To do this, we \emph{guide} the classifier's attention towards the area highlighted by the saliency from the language specification. Finally, the visual classifier still needs to solve the more fine-grained task, after obtaining the high-level attention guidance. It thus retains some flexibility, e.g. it may even attend to some useful (non-harmful) context. In practice, we use the recent powerful CLIP~\cite{radford2021learning} model to ground textual information into images. We leverage the ``Right for the Right Reasons'' method~\cite{ross2017right} to enforce that the classifier indeed attends according to the given guidance. With this approach, we can incorporate language specification via an auxiliary loss during training, and thus the \emph{vision-language model is not needed during inference}.

We show how \method{} can assist in training on data with explicit and implicit bias. On the synthetic Waterbirds dataset~\cite{sagawa2019distributionally} which contains a known, explicit bias (the image backgrounds), our method is able to achieve  $\sim$2$-$7\% per-group accuracy improvements over baselines, including a model which uses an unsupervised attention mechanism instead of guidance from language. \method{} also shows a 15\% improvement on the worst-group accuracy in the challenging scenario where class labels are perfectly correlated with the distracting backgrounds (\cref{sec:wbirds}).  For implicit bias, where training and test distributions differ in unknown ways, we see that \method{} achieves $\sim$41$-$45\% relative improvements on fairness metrics for apparent gender recognition (\cref{sec:coco}). We also show a 2\% accuracy improvement on a red-meat classification task from a subset of Food-101~\cite{food_101}, where an implicit bias emerges from noisy training labels (\cref{sec:meat}). Lastly, we demonstrate that the quality of classifiers' explanations improves with the given advice (12.8\% improvement in Pointing Game~\cite{zhang2018top} accuracy, described in \cref{sec:pg}). We provide our code and datasets to reproduce our experiments.

\section{Related Work}
\label{sec:related}

\textbf{Addressing bias with instance annotations.}
Most prior works that address bias in visual classifiers assume that some form of instance annotation is available. Some rely on expensive spatial annotations, such as object masks~\cite{hendricks2018women,li2018tell,rieger2020interpretations} or bounding boxes~\cite{choi2019can}. Hendricks~\etal\cite{hendricks2018women} address the image captioning task, where they want to reduce bias amplification and ensure a fair outcome for male and female genders by using person masks at training time. Others use slightly less expensive image-level annotations of the biased feature~\cite{adeli2021representation,kim2019learning,sagawa2019distributionally,singh2020don,wang2019balanced}. %
In contrast, in this work we do not assume that instance-level bias information is available. Instead, we rely on automatically generating attention guidance with readily-available language specification.

\textbf{Addressing bias without instance annotations.}
Several works address bias without explicitly relying on instance-level bias annotations~\cite{clark2020learning,nam2020learning,utama2020towards,liu2021just}. Clark~\etal~\cite{clark2020learning} train an ensemble of low and high capacity models, forcing them to be conditionally independent, with the hope that the low capacity model will learn bias features, and the high capacity model will then learn the task-relevant features. 
Nam~\etal~\cite{nam2020learning} also train two models, one ``biased'' and the other ``unbiased'', by amplifying samples ``aligned'' with the bias for the first model (or easier to learn at the early stages), while amplifying the more difficult samples for the second model (where the first one fails). 
We view this line of research as complementary to our effort, and envision potentially combining these ideas with ours.

\textbf{Language as information for visual tasks.}
We draw inspiration from prior work that leverages language in vision learning systems. Incorporating language in the zero/few-shot setting has been widely explored. Embedding language from class names or descriptions to obtain class ``prototypes'' is common in zero-shot learning, when no visual samples of the class are available~\cite{elhoseiny2013write,qiao2016less,le2020using,frome2013devise,radford2021learning}. Several works also aim to learn classes using their semantic attributes for better knowledge transfer~\cite{lampert_zs, mall2021field, xu2020attribute}. Mu~\etal~\cite{mu2020shaping} use image captions for regularizing few-shot representations to hold semantically meaningful information.
Outside of zero/few-shot learning, Kim~\etal~\cite{kim2020advisable} incorporate language advice into an autonomous driving controller, leading to a better performing and more explainable model. 
Rupprecht~\etal~\cite{rupprecht2018guide} use language interactively to improve a pretrained CNN during inference time on semantic segmentation tasks.
Ling~\etal~\cite{ling2017teaching} use language feedback to improve an image captioning model.
To the best of our knowledge, no works have explored using language specification to improve visual attention in biased scenarios.

\textbf{Information grounding with vision-language models.}
One of the key components of our approach is to leverage an off-the-shelf vision-and-language model to ground textual information into an image. There is a large body of work on visual grounding, where the models are trained to localize textual expressions in an image with a bounding box~\cite{plummer2015flickr30k,rohrbach2016grounding} or a segmentation mask~\cite{hu2016segmentation}. Unfortunately, these methods are constrained by the cost of providing these extra labels for the training set. 
Others can handle more open-ended queries, but the size of the available training data is small as they require localization supervision which is costly to obtain, limiting the general application of these methods~\cite{kazemzadeh2014referitgame,plummer2015flickr30k}. A recent vision-and-language model CLIP~\cite{radford2021learning} has demonstrated state-of-the-art image-text retrieval capabilities. 
CLIP is trained on 400M image-caption pairs sourced from the Web, making it a powerful general-purpose representation. 
We use CLIP and obtain grounding information with the help of salience visualization techniques~\cite{selvaraju2017grad}.

\textbf{Supervising spatial attention in visual classifiers.}
Another important component of our method is guiding the spatial attention within the visual classifier away from the spurious features. Several prior works have explored supervising spatial attention for, e.g., preventing catastrophic forgetting~\cite{ebrahimi2020remembering}, fine-grained recognition~\cite{fukui2019attention}, domain transfer~\cite{zunino2020explainable} and generation of faithful explanations~\cite{ross2017right}. Specifically, the Right for Right Reasons approach~\cite{ross2017right} penalizes large input gradients in regions that are not allowed based on the user-defined ``right reasons''. We leverage this method to guide the classifier's attention towards the evidence pointed out by the language specification.

\section{
Guiding Attention with Language
}
\label{sec:method}

\begin{figure*}[t]
  \centering
  \includegraphics[width=.8\linewidth]{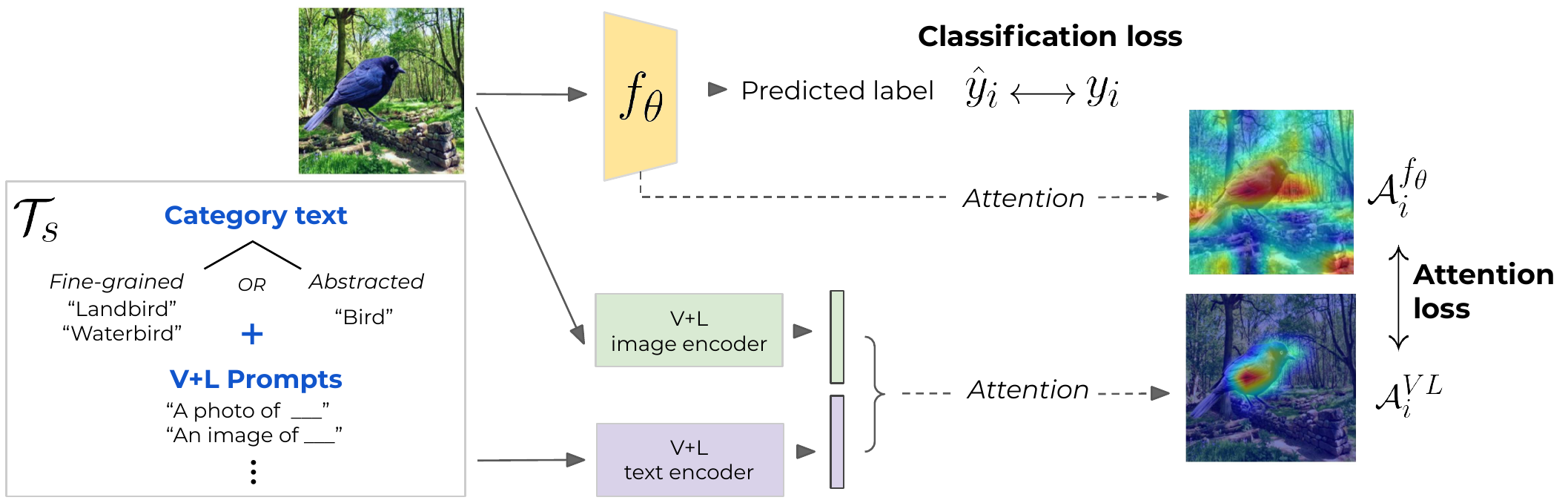}
  \caption{\textbf{\method{} overview.} Our framework consists of three parts. First, we create a language specification $\mathcal{T}_s$ based on provided class names or a description of the task.
  Next, for every training image $x_i$, we use a pretrained vision and language model to ground the textual information into an image, in the form of an attention map $A^{VL}_i$. Finally, when we train the classifier $f_{\theta}$, we incorporate $A^{VL}_i$ as attention supervision. This encourages $f_{\theta}$ to align its attention $A^{f_{\theta}}_i$with task-relevant concepts, and away from distractors.}
  \label{fig:method}
   \vspace{-2mm}
\end{figure*}

In the following we outline \method{}, our framework for incorporating language specification to guide a visual classifier; \cref{fig:method} provides an overview of our approach.

\textbf{Problem Definition.} 
In this work, we consider the learning problem in which we are given an image classification dataset $\left\{x_i, y_i\right\}_{i=1}^n$ for a prediction task $\mathcal{T}$ with $C$ classes. Additionally, we assume we have a corresponding natural language specification $\mathcal{T}_s$ of the task or language descriptions of each class within the task $\mathcal{T}_s^c$.
We also assume each image $x_i \in \mathbb{R}^{h\times w\times d}$ may contain a region of pixels that is irrelevant to $\mathcal{T}$, yet strongly correlated with $y_i$.

To model the distinction between the relevant and spuriously correlated pixels, 
we introduce a latent binary mask $\z^{\mathcal{T}}_i\in \{0,1\}^{h\times w}$ for each image $x_i$, which
encodes the relevance of each pixel to the task $\mathcal{T}$. 
That is, if $\z^{\mathcal{T}}_{i,(u,v)} = 1$, then the value of pixel $x_{i,(u,v)}$ is informative for task $\mathcal{T}$ (and 0 if otherwise).
Note that $\z^{\mathcal{T}}_i$ is dependent on the prediction task. However, for notational convenience, we will omit $\mathcal{T}$ from $\z^\mathcal{T}_i$ in the following.

Next, consider an image classification model $f_{\theta}$ with parameters $\theta$. Our goal is to learn an optimal classifier $f_{\theta^*}$, which outputs predictions $\hat y$ that rely only on task-relevant features (where $\z_i=1$).
As $\z_i$ is unobserved, we cannot learn $f_{\theta^*}$ by simply masking images according to locations of relevant features. Instead, we want to estimate a probability map over $\z$, where each entry $x_{i,(u,v)}$ corresponds to the probability that pixel $x_{i,(u,v)}$ is relevant to $\mathcal{T}$. 

Given this setup, our framework is three-fold: first, we create the high-level natural language specification $\mathcal{T}_s$ describing the semantic concepts relevant to $\mathcal{T}$. This is based on provided class names (e.g. ``landbird'') or description of the task (e.g. ``bird species classification'').
We then use a pretrained vision-language model and a spatial attention function to compute an estimate of the task attention $\z$ for each image w.r.t. $\mathcal{T}_s$. 
Lastly, we use these estimates to supervise the spatial attention of $f_{\theta}$, guiding it towards task-relevant features and away from unwanted biases.

\textbf{Language specification.} We assume access to natural language class names or a description of the task $\mathcal{T}$, but not necessarily access to what biases exist in the data. We argue that this is a safe assumption -- in most real-world classification tasks, it is expected that a user has knowledge of what the categories are or what the task means. We then use the provided natural language to create $\mathcal{T}_s$ -- words or phrases which are compatible with the choice of pretrained vision-language model, described below. For example, we preface task-relevant phrases with ``a photo of'' or ``an image of'' for compatibility with CLIP~\cite{radford2021learning}. The language specification can be the same for each instance, or it can be class-specific by using the labels provided during training. Note that $\mathcal{T}_s$ is created once, prior to the training of $f_{\theta}$, and does not require annotation of each image individually, allowing our framework to easily scale to large datasets.

\textbf{Generating an estimate of $\z$ from language specification.} Consider a pretrained multimodal vision-and-language model $VL$, which has a joint understanding of image features and language phrases that correspond to them. For example, $VL$ can be an image captioning or visual grounding model, or a model trained at scale with joint image-text supervision, such as OSCAR~\cite{li2020oscar}, VinVL~\cite{zhang2021vinvl}, or CLIP~\cite{radford2021learning}, the latter of which we use in our experiments.

For every image $x_i$ in the training dataset, we precompute a spatial attention map $\mathcal{A}^{VL}_i = Att^{VL}(\mathcal{T}_s^{y_i},x_i)$, with $\mathcal{A}^{VL}_i \in \mathbb{R}^{h\times w}$. This serves as a probability map over $\z_i$, where the attention value at location $(u,v)$ estimates the likelihood that pixel $x_{i,(u,v)}$ is a task-relevant feature. %
The quality of $\mathcal{A}^{VL}$ as an estimate of $\z$ depends on the ability of the pretrained vision and language model to ground text phrases in visual features. %
However, proper grounding within vision-and-language models is a research question on its own~\cite{liu2017attention,rohrbach2018object}. Luckily, 
recent work on large-scale image-language pretraining has led to promising improvements~\cite{li2020oscar, zhang2021vinvl,  radford2021learning}. Here, we use the saliency method GradCAM~\cite{selvaraju2017grad} to obtain reasonable attention maps.

Generating an estimate of the true task attention $\z$ in this manner provides an automatic method for localizing per-instance, task-relevant features according to user specification $\mathcal{T}_s$. It requires only a high-level description of which semantic concepts are relevant to a task, which we view as a valid assumption for a user of a machine learning system.

\textbf{Guiding the classifier with spatial attention.} Next, for each image $x_i$, our objective is to guide the spatial attention of the classifier $f_{\theta}$ away from spurious correlations and towards task-relevant features. To do so, we would like to supervise the spatial attention of $f_{\theta}$ with the $\mathcal{T}_s$ attention maps $\mathcal{A}^{VL}_i$ computed in the previous step of our framework.
This requires a function $Att^{f_{\theta}}(x_i,y_i)$, which computes a differentiable attention map $\mathcal{A}^{f_{\theta}}_i$. The attention map specifies spatial locations in $x_i$ that were relevant to the prediction $\hat y_i$. %

We supervise the classifier's attention for each training image $x_i$ by computing a loss $\mathcal{L}_{att}$ between $\mathcal{A}^{VL}_i$ and $\mathcal{A}^{f_{\theta}}_i$. The final training loss $\mathcal{L}(\theta, X, y, \mathcal{A}^{VL})$ for a batch of training images $X$ with $m$ samples is given as:
\begin{equation}
\label{eq:loss}
    \mathcal{L}(\theta, X, y, \mathcal{A}^{VL}) = -\frac{1}{m} \sum\limits_{i=1}^m y_i \cdot log(\hat y_i) + \lambda \mathcal{L}_{att}(\mathcal{A}^{VL}_i, \mathcal{A}^{f_{\theta}}_i)
\end{equation}
where $\lambda$ is a hyperparameter that controls the strength of the attention supervision.

Our proposed framework does not require an architectural change to the classifier $f_{\theta}$, and only incorporates language-guided spatial attention as an auxiliary loss term in \cref{eq:loss} during training time. Therefore, 
our framework requires no additional knowledge at test time.

\subsection{Model Design Choices}\label{sec:implement}

\textbf{Vision-Language model.} We use the CLIP (Constrastive Language-Image Pre-training model)~\cite{radford2021learning} as our multimodal $VL$ model. CLIP is trained on 400M image-caption pairs $(x_{text}, x_{image})$ sourced from the Web. It consists of two encoders for mapping $x_{text}$ and $x_{image}$ into a shared embedding space. The contrastive objective encourages image and text from the same pair to be close in the embedding space (as measured by cosine distance), and image and text from different pairs to be pushed apart.%

\textbf{Generating attention.} For the language specification $\mathcal{T}_s$, we define a set of CLIP-style prompts. These are framed as short sentence descriptions, such as ``an image of \textit{X}'' or ``a photo with \textit{X},'' for the word or phrase \textit{X} that describes task-relevant concepts. We generate multiple such prompts for each task and later combine (via average or max) the corresponding attention maps for each image, which serves as our estimate for $\z$. Once the prompts are defined, they are embedded with CLIP's text encoder into the shared image-text latent space. For embedding images, we use the image encoder of CLIP with the ResNet50 backbone provided by Radford~\etal~\cite{radford2021learning}. For the attention function $Att^{VL}(\mathcal{T}_s, x_i)$, we use the saliency method GradCAM~\cite{selvaraju2017grad} between the image-text similarity score and the feature maps after the last convolutional block in the image encoder.

\textbf{Attention incorporation.} For supervising the classifier's attention, we adapt the framework Right for the Right Reasons, or RRR~\cite{ross2017right}. The original goal of RRR was to provide the correct explanation for each sample in addition to the correct prediction. This aligns well with our goal to prevent a model from learning unwanted feature correlations. 
In RRR, a user provides per-image binary masks of regions that are \textit{irrelevant} to the task. It then penalizes the input gradients in those regions (the gradient of the output $y$ with respect to the input $x$). In our work, the  attention maps $\mathcal{A}^{VL}_i$ specify \textit{relevant} regions to the task. Therefore, we take $1-\mathcal{A}^{VL}_i$ to specify \textit{irrelevant} regions, and we compute the L1 loss between this and the input gradient. We normalize $\mathcal{A}^{VL}_i$ to contain values between 0 and 1 (instead of using a binary mask as in the original RRR method), as our intention is to estimate a probability map over the true task attention $\z$.  %

\textbf{Loss function.} 
We apply GradCAM~\cite{selvaraju2017grad} to our chosen $VL$ model (CLIP with a ResNet50 backbone), to 
provide $\mathcal{A}^{VL}_i$, 
the input gradients for a ResNet50 model pretrained on ImageNet as $\mathcal{A}^{f_{\theta}}_i = \frac{dy}{dX_i}$, and the RRR-based loss for $\mathcal{L}_{att}$. Thus, our loss function used in the experiments is:
\begin{equation}\label{eq:rrr}
\begin{aligned}
\mathcal{L}(\theta, X, y, \mathcal{A}^{VL}) = {\overbrace{\textstyle {-\frac{1}{m} \sum\limits_{i=1}^m y_i \cdot log(\hat y_i)}}^{\mathclap{\text{Classification loss}}}}  \\
+ \: \: {\underbrace{\textstyle \frac{\lambda}{m} \sum\limits_{i=1}^m \left|\frac{dy}{dX_i}(1-\mathcal{A}^{VL}_i) \right|}_{\mathclap{\text{Attention loss}}}}
\end{aligned}
\end{equation}

Our proposed framework is not restricted to a specific choice of pretrained model $VL$, classifier $f_{\theta}$, mechanism of generating attention maps  $\mathcal{A}^{VL}$ and $\mathcal{A}^{f_{\theta}}$, and attention loss function $\mathcal{L}_{att}$. \method{} in the following section refers to the particular choices described above. We include ablations in Sec.~\ref{sec:ablations} for other choices of $VL$, $\mathcal{A}^{f_{\theta}}$, and $\mathcal{L}_{att}$.

\section{Experiments}
\label{sec:experiments}

\textbf{Training.} We use a ResNet50~\cite{he2016deep} backbone pretrained on ImageNet for all classification models, with an input image resolution of $(224,224)$. The GradCAM attention maps from CLIP are of size $(7,7)$, which is the spatial resolution of the activations from the last convolutional block. We resize them up to the input resolution before computing the L1 loss. 
All error bars show standard deviation across 10 trials.
We report further details on training parameters (such as the loss weight $\lambda$) and hyperparameter sweeps in the Appendix.

\textbf{Baselines.} We compare our work with several baselines that  do not require per-instance knowledge of bias features. All baselines use the same ResNet50 backbone for consistency. \textit{Vanilla} is trained in the same manner as $f_{\theta}$ in our framework, except without the attention loss $\mathcal{L}_{att}$. \textit{UpWeight} is the same as Vanilla, except it uses class labels to address class imbalance. It computes a weighted average of per-sample cross entropy. The weights are inversely proportional to the frequency of the sample's class in the training data, assigning a weight of 1 to the class with fewest samples. \textit{Attention Branch Network}, or \textit{ABN}~\cite{fukui2019attention}, learns a feed-forward attention map before the last convolutional block of ResNet50 and element-wise multiples it with the activations, which is added back into the activations before passing to the rest of the model. It also adds an additional cross-entropy loss term based on features in the attention branch, to encourage the spatial attention to be class-specific\footnote{We also experimented with supervising the attention of ABN with language specification. However, it under-performed the current formulation, and we include it in ablations in Section~\ref{sec:ablations}.}. We include tabular results of plots in the Appendix.

\textbf{Visualizations.} For all visualizations, the attention from language specification is generated with GradCAM (as in Sec.~\ref{sec:implement}), and classifier attentions are generated with the black-box saliency method RISE~\cite{petsiuk2018rise}. More examples of attention for each dataset are in the Appendix.
\subsection{Datasets}\label{sec:datasets}

We evaluate our approach on datasets with explicit and implicit bias. Additional details are in the Appendix, including dataset size and creation. The license, PII, and consent details of each dataset are in the respective papers.

In the \emph{explicit} bias setting, the distractor feature can be clearly defined and (potentially) labeled. 
We experiment with the synthetic Waterbirds dataset~\cite{sagawa2019distributionally}, where bias is easy to control. Specifically, the images of birds from the CUB dataset~\cite{wah2011caltech} are divided in two classes, landbirds and waterbirds. Next, birds are segmented out and pasted onto random land or water backgrounds from the Places dataset~\cite{zhou2017places}. During training, most waterbirds appear on water backgrounds and landbirds on land backgrounds, while in validation/test sets each class has an equal number of samples on land and water backgrounds. We consider two scenarios, one in which there is a small fraction of samples ($5\%$) in the training data that go against the bias (\textbf{\emph{Waterbirds-95\%}}) and a more challenging one, where the bias and labels are perfectly correlated during training (\textbf{\emph{Waterbirds-100\%}}).

The Food-101 dataset~\cite{food_101} presents a case of \emph{implicit} bias, as it was intentionally created such that the training images were not cleaned -- for example, the images contain noise in the form of incorrect labels, bright colors, and visual confusion. Certain other foods appear more frequently with some classes than the others (e.g. sauce appears more often with baby back ribs than with steak). The evaluation set, on the other hand, was more thoroughly cleaned. %
We construct a 5-way~\textbf{\emph{Red Meat}} classification task between baby back ribs, filet mignon, pork chop, prime rib, and steak.

We present a second dataset  with implicit bias, \textbf{\emph{MSCOCO-ApparentGender}}, which is constructed based on MSCOCO~Captions~\cite{chen2015microsoft} and prior work~\cite{hendricks2018women,zhao2017men}. In this dataset, apparent gender labels are defined based on the people's outward appearance as reflected in image captions. As defined in \cite{hendricks2018women}, when discussing people in captions, there are three options: ``Man'', ``Woman'' or a gender-neutral term, e.g. ``Person''. To follow that, we consider a three-way classification task for apparent gender, using the provided captions to generate labels for the classes ``Man'', ``Woman'', and ``Person'' (the latter when the annotators did not use gendered words in the captions). There are different types of spurious correlations in this dataset, e.g. women appearing in some environments more often than men, or a distractor object co-occurring with men but not with women, etc. 

\begin{figure}
     \centering
     \begin{subfigure}[t]{0.4\textwidth}
         \centering
         \includegraphics[width=\textwidth]{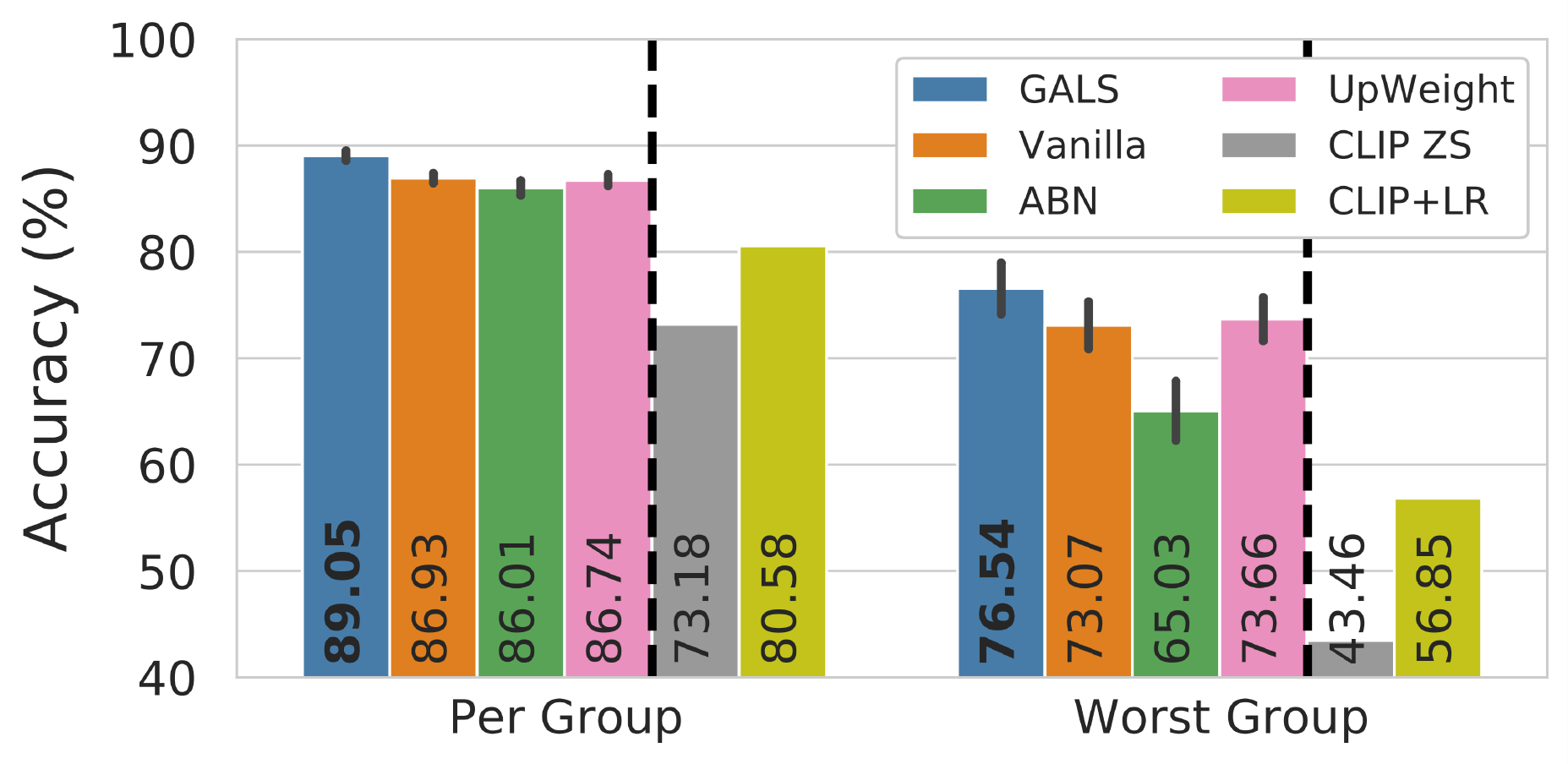}
         \caption{Waterbirds 95\%}
         \label{fig:waterbirds_95}
     \end{subfigure}
     \begin{subfigure}[t]{0.4\textwidth}
         \centering
         \includegraphics[width=\textwidth]{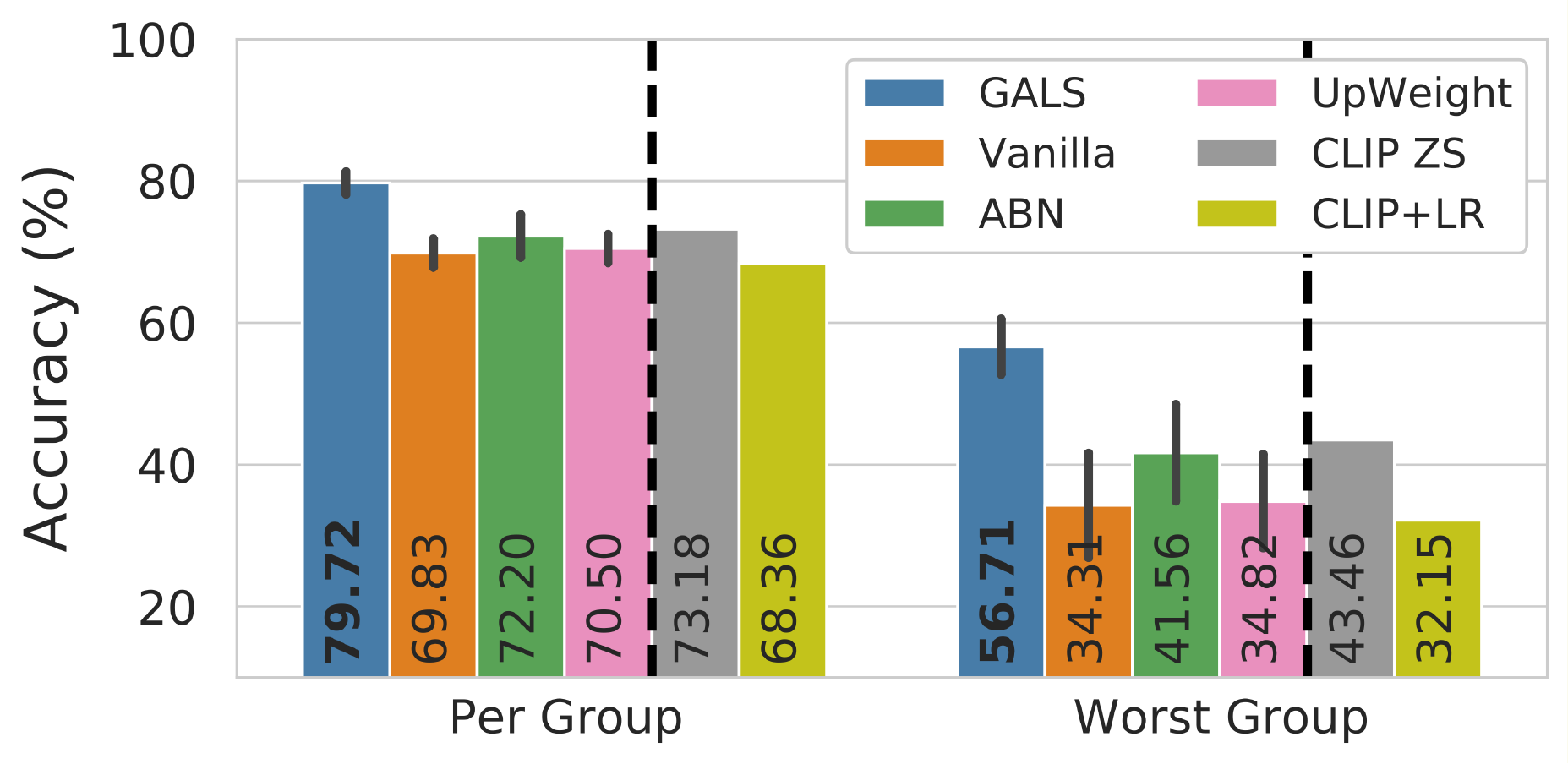}
         \caption{Waterbirds 100\%}
         \label{fig:waterbirds_100}
     \end{subfigure}
        \caption{\textbf{\emph{Waterbirds}.} Test accuracy on \emph{Waterbirds-95\%} and  \emph{Waterbirds-100\%} datasets. Incorporating language specification results in a higher accuracy than all other baselines, including zero-shot CLIP and CLIP finetuned with logistic regression.}
        \label{fig:waterbirds}
         \vspace{-3mm}
\end{figure}

\subsection{Explicit bias on Waterbirds}
\label{sec:wbirds}

Since the Waterbirds dataset is constructed to encourage the model to pay attention to the background and not the bird, high-level language specification should give direction to attend to the bird, leaving the fine-grained discriminative image features up to the classifier to discover. Specifically, we generate attention from two CLIP prompts, to reduce noise -- ``an image of a bird'' and ``a photo of a bird.'' We average together these per-sample attentions to obtain $\mathcal{A}^{VL}_i$\footnote{In rare cases, the attention for a single prompt would be all-zero. Instead of averaging, we use the non-zero attention from the second prompt.}.

Following~\cite{sagawa2019distributionally}, we present test accuracy \emph{per-group}, in which accuracy is weighted equally over the groups (specific combinations of class label and background, i.e. landbirds on land, landbirds on water, waterbirds on land, and waterbirds on water), and the \emph{worst-group} accuracy. We are particularly interested in the worst-group (usually waterbird on land) performance, which suffers the most when a model makes use of the spurious background correlations.

The concepts ``landbird'' and ``waterbird'' are rare with respect to concepts that can be learned from the Web, as would often be the case in new, real-world classification tasks. To illustrate that large-scale models like CLIP may lack fine-grained task-specific knowledge, we compare our method to zero-shot CLIP, as well as logistic regression trained on top of CLIP image-encoder features (following~\cite{radford2021learning}). We find that CLIP often underperforms even the Vanilla baseline, demonstrating the value of taking the ``best of both worlds'' by combining large-scale multimodal model attention with CNNs on biased datasets with unfamiliar concepts. %

\textbf{Waterbirds-95\%:} 
As shown in Fig.~\ref{fig:waterbirds_95}, our method outperforms all baselines on both per-group and worst-group accuracy. The strong bias in the data is evident when considering the worst-group accuracy, which drops the Vanilla performance by about 14\%. Our model drives up the worst-group performance by 2.88\% from the next-closest baseline of class weighting, without sacrificing per-group accuracy.

\textbf{Waterbirds-100\%:} Because the class label and background are perfectly correlated, the performance of a classifier without any additional task information depends on whether it is easier to capture the true or bias signal. Surprisingly, the unsupervised attention mechanism in ABN provides $\sim$7\% boost in worst-group performance as compared to upweighting by the class label. Our model improves on this, leading to a 15.15\% improvement over ABN.
\subsubsection{Using language specification to change the task}
Since the ``landbird'' and ``waterbird'' labels in the \emph{Waterbirds-100\%} training set are perfectly correlated with land and water backgrounds, we can easily redefine the labels to create a background classification task.
We investigate whether we can use language specification to choose which hypothesis a model learns during train time: the ``bird'' or the ``background'' task. To study this, we keep the training set the same, yet update the validation and test labels to reflect background classification. We use the phrases ``nature scene'', ``outdoor scene'', and ``landscape'', preceded with ``a photo of'' and ``an image of'' as in our other experiments. We ensemble the attention maps by taking the max value for each pixel. A vanilla ResNet50 baseline achieves $86.75\%$ per-group accuracy on the test set, with a worst-group accuracy of $72.90\%$. Impressively, our method outperforms this baseline by  $\textbf{2.22\%}$ and $\textbf{7.32\%}$ on per-group and worst-group accuracy respectively, demonstrating the flexibility of language specification to select the desired training signal.

\begin{table}[b]
\resizebox{1\linewidth}{!}
{
  \begin{tabular}{lrrrr}
    \toprule
       & \textbf{\method{}} & \textbf{Vanilla} & \textbf{ABN}~\cite{fukui2019attention} \\
    \midrule
    Accuracy (\%) & \textbf{71.20} $\pm$ \textbf{0.84} & 67.39 $\pm$ 0.88 & 69.44 $\pm$ 1.12 \\
    \bottomrule
  \end{tabular}
  }
\caption{\textbf{\emph{Red Meat}.} Test accuracy of our method, vanilla, and ABN for \emph{Red Meat} Classification (a subset of the Food-101 dataset).}
\label{tab:food}
\end{table}

\begin{table*}[h]

  \centering
  \small

  \begin{tabular}{lrrrrrrrr}
    \toprule
     &   \multicolumn{3}{c}{\textbf{Man}}    & \multicolumn{3}{c}{\textbf{Woman}}                   \\
    \cmidrule(r){2-4} \cmidrule(r){5-7}
    \textbf{Method}   & \emph{Man} & \emph{Woman} & \emph{Other} & \emph{Woman} & \emph{Man} & \emph{Other} & \textbf{Ratio $\Delta$} & \makecell{\textbf{Outcome}\\ \textbf{Divergence}}\\
    \midrule
    Vanilla &  \underline{83.60} &	\underline{6.20} & 10.20 &	66.80 &	28.60 &	4.60  & 0.349 & 0.071\\
    ABN~\cite{fukui2019attention} & \textbf{84.80} & \textbf{4.60} & 10.60 & \underline{68.80} & \underline{25.40} & 5.80 & 0.339 & 0.068\\
    UpWeight &  80.20 &	11.20 &	8.60 &	68.00 &	28.60 &	3.40 & \underline{0.272} & \underline{0.040}\\
    \method{} & 79.80 &	11.80 &	8.40 &	\textbf{74.20} & \textbf{22.60} &	3.20 & \textbf{0.160} & \textbf{0.022}\\
    \bottomrule
  \end{tabular}
  \caption{\textbf{\emph{MSCOCO-ApparentGender}.} Performance of our approach and the baselines on \emph{MSCOCO-ApparentGender} test set. The best result in each column is \textbf{bold}, and second-best is \underline{underlined}.
   }
   \label{tab:gender}
    \vspace{-2mm}
\end{table*}

\begin{figure}[t]
  \centering
  \includegraphics[width=0.85\linewidth]{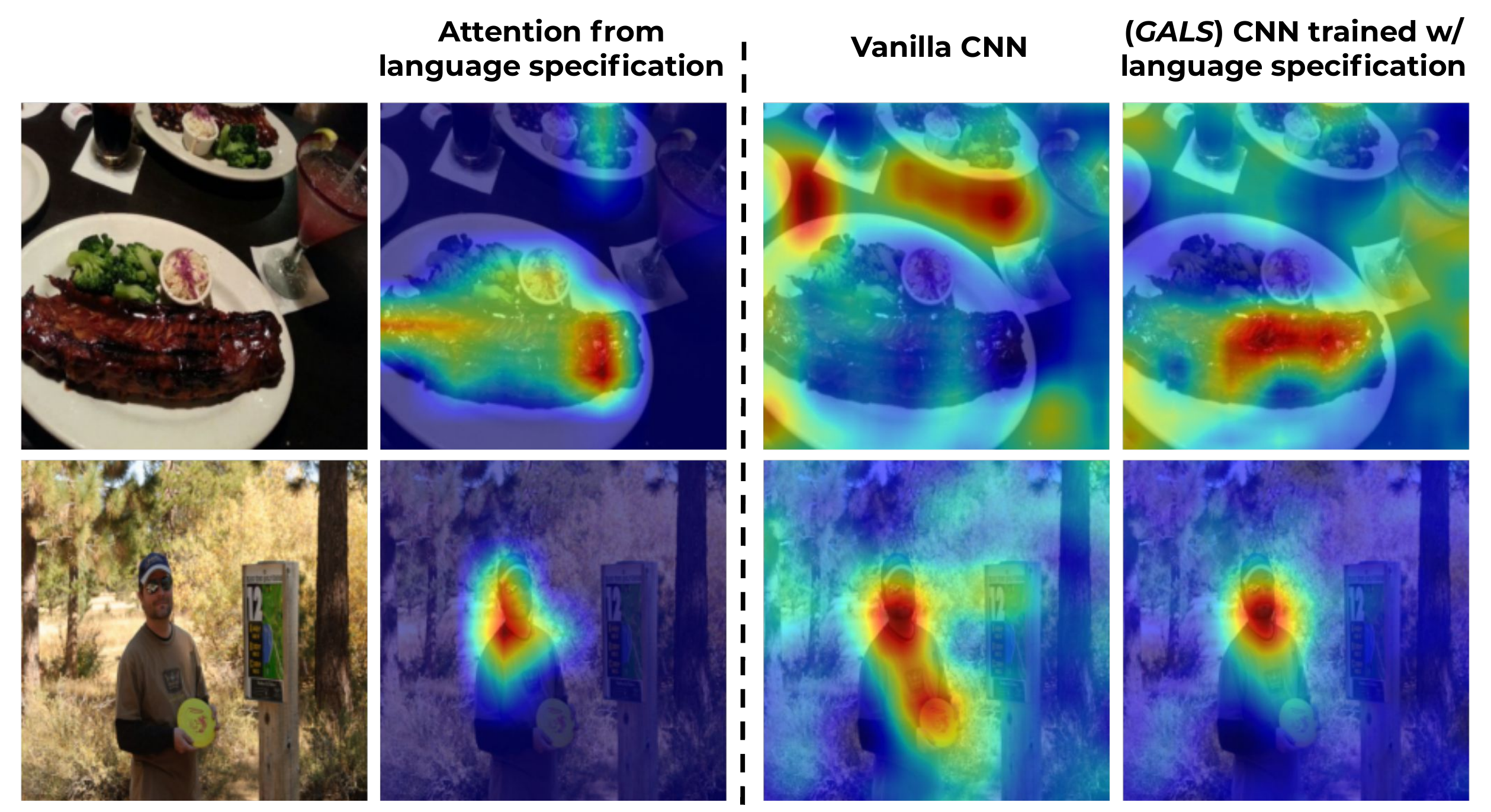}
  \caption{\textbf{Qualitative results for implicit bias.} Sample attention on \emph{Red Meat} (top) and \emph{MSCOCO-ApparentGender} (bottom). On these datasets %
  the vanilla classifier may attend to non-task-relevant features due to implicit biases or noise. When we ground relevant features with language specification, we are able to move the classifier's attention away from the distractors.}
  \label{fig:gender_food_vis}
   \vspace{-2mm}
\end{figure}

\subsection{Red Meat Classification with Noisy Data}\label{sec:meat}

Along with assisting in removing explicit contextual bias in datasets, in this experiment we will show how our approach can improve the learning process on implicit bias caused by noisy data. We train on 5 balanced classes from the Food-101 dataset pertaining to red meat, as discussed earlier. We generate attention from the CLIP prompts ``an image of meat'' and ``a photo of meat''.
Our results displayed in Table~\ref{tab:food} and visualized in Fig.~\ref{fig:gender_food_vis} (top) show that our method is able to outperform the ABN model by $\sim2\%$ overall.

\subsection{Implicit bias on MSCOCO-ApparentGender}\label{sec:coco}

Next, we discuss how our approach performs in another implicit bias scenario on the \emph{MSCOCO-ApparentGender} dataset. We follow the evaluation protocol from \cite{hendricks2018women} and generate attention from the CLIP prompts ``an image of a person'' and ``a photo of a person''. Table~\ref{tab:gender} summarizes the quantitative results and Fig.~\ref{fig:gender_food_vis} (bottom) displays a qualitative example of the attention maps. For each ``Man'' / ``Woman'' sample we separately report the \% of the time they have been classified as a \emph{Man}, \emph{Woman}, or \emph{Other}. We penalize gender misclassification, but do not penalize if the ``Person'' class was predicted. In this task, we care about several aspects. (1) The training data is imbalanced (with more men than women in it), thus we aim to reduce bias amplification at test time~\cite{hendricks2018women}. The metric ``Ratio Delta'' measures how close the predicted men/women ratio is to the true one (which is equal to $1.0$), i.e. lower is better. Our approach performs the best in this metric. (2) We also aim to ensure an equal outcome for both men and women. In practice, we see that men tend to be recognized more accurately than women, as seen from the higher \textbf{Man}/\emph{Man} values than the \textbf{Woman}/\emph{Woman} values (e.g., the Vanilla baseline achieves 83.6\% and 66.8\% accuracy, respectively). As we see, women often get misclassified as men (22$-$28\% across methods). The ``Outcome Divergence'' metric measures Jensen-Shannon divergence~\cite{lin1991divergence} between the two sets of scores across the two classes, i.e. lower is better~\cite{hendricks2018women}. Again, our approach achieves the lowest outcome divergence, demonstrating the most fair behavior across all the compared methods. %

\begin{table*}[h]
\centering
        \begin{subtable}[h]{0.44\textwidth}
        \centering
        \resizebox{1.1\linewidth}{!}
        {
        \begin{tabular}{llrrr}
            \toprule
              &  & \multicolumn{3}{c}{\textbf{Validation Accuracy}}                   \\
            \cmidrule(r){3-5}
             \makecell{ \textbf{Classifier} \\ \textbf{Attention Method}} & \makecell{\textbf{Language}\\  \textbf{Attention Source} }  & \textbf{Per Class} & \textbf{Landbird} & \textbf{Waterbird} \\
            \midrule
              ABN & CLIP ViT  &	86.93 & 90.78 & 83.08 \\
              ABN & CLIP R50  & 86.10 & 90.25 & 81.95\\	
              GradCAM  & CLIP ViT  & 87.20 & \underline{91.32} & 83.08 \\
              GradCAM & CLIP R50  	 & 84.44 & 89.92 & 78.95 \\
              RRR  & CLIP ViT  & \textbf{88.25} & \textbf{92.28} & \underline{84.21}	\\
              \midrule
             RRR & CLIP R50  & \underline{87.26} & 89.17 & \textbf{85.34}\\
            \bottomrule
          \end{tabular}
        }
        \caption{\emph{Waterbirds-95\%}}
        \label{tab:waterbirds_95_design}
     \end{subtable}
    \hspace{1cm}
    \begin{subtable}[h]{0.46\textwidth}
        \centering
        \resizebox{1.1\linewidth}{!}
        {
                \begin{tabular}{lrrrrrrrrr}
            \toprule
             &   & \multicolumn{3}{c}{\textbf{Man}}    & \multicolumn{3}{c}{\textbf{Woman}}                   \\
            \cmidrule(r){3-5} \cmidrule(r){6-8}
          \makecell{ \textbf{Cls.} \\ \textbf{Att.} \\ \textbf{Method}} & \makecell{\textbf{Lang.}\\  \textbf{Att.} \\ \textbf{Source} } & \emph{Man} & \emph{Woman} & \emph{Other} & \emph{Woman} & \emph{Man} & \emph{Other} & \textbf{R$\Delta$} & \textbf{OD}\\
            \midrule
            ABN & CLIP ViT & 84.40 & 10.60 & 5.00 & 68.40 & 29.40  & 2.20 & \underline{0.306} & \textbf{0.274}\\
            ABN & CLIP R50 & \textbf{90.60} & \textbf{5.40} & 4.00 &   60.40 & 37.60 & 1.80 & 0.485 & \underline{0.280}\\
            GradCAM & CLIP ViT & 85.80 & 7.60 & 6.60 & \underline{70.20} & \underline{27.00}  & 2.80 & 0.310 & 0.331 \\
            GradCAM & CLIP R50 & 83.40 & \underline{7.40} & 9.20 & 66.20 & 29.80  & 4.00 & 0.311 & 0.298\\
            RRR & CLIP ViT & \underline{87.00} & 8.40 & 4.60 & 68.60 & 29.80  & 1.60 & 0.341 & 0.305 \\
            \midrule
            RRR & CLIP R50 & 82.20 & 10.60 & 7.20 & \textbf{72.20} & \textbf{26.00}  & 1.80 & \textbf{0.235} & 0.309\\
            \bottomrule
          \end{tabular}
        }
      \caption{\emph{MSCOCO-ApparentGender}}
      \label{tab:coco_design}
    \end{subtable}
    \caption{Comparison of different classifier attention methods and language attention sources on the (a) \emph{Waterbirds 95\%} and (b) \emph{MSCOCO-ApparentGender} validation set.
    In (a), we report class instead of group scores, as we do not assume access to group labels at validation.
    The method indicated as ``\method{}'' in Section 4 is placed at the bottom.}
     \label{tab:design_choices}
\end{table*}

\subsection{Attention Evaluation}\label{sec:pg}

\begin{figure}[t]
  \centering
  \includegraphics[width=0.95\linewidth]{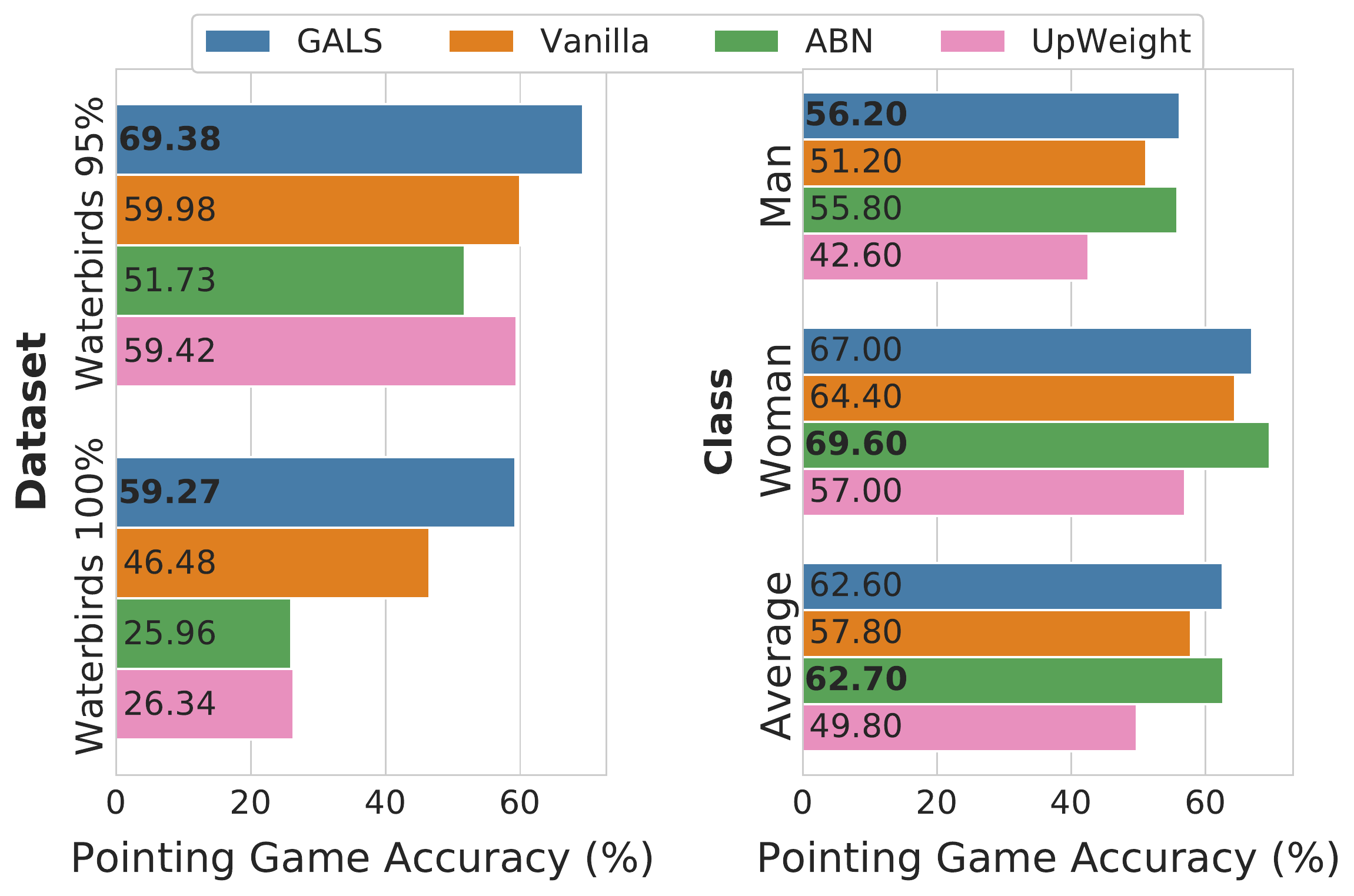}
  \caption{\textbf{Pointing game.} Pointing Game experiment~\cite{zhang2018top} on \emph{left.} Variants of Waterbirds datasets and \emph{right.} MSCOCO-ApparentGender. We test whether the peak value of a black-box model explanation, generated with RISE~\cite{petsiuk2018rise}, falls inside the segmentation label of the salient object.}
  \label{fig:pg}   
  \vspace{-5mm}
\end{figure}

We evaluate the quality of our model explanations to determine if language specification makes the model right for the right reasons in addition to improving accuracy. 
To do so, we use the Pointing Game~\cite{zhang2018top}, a common evaluation for model explanations. For each input $x_i$, the Pointing Game (PG) requires a corresponding model explanation $a_i$ and binary mask $\z_i$, both of the same dimensions as $x_i$. Recall that $\z_i$ indicates the task-relevant pixels in an image. A model passes the PG on sample $i$ if the maximum value of its explanation $a_i$ falls inside $\z$. In other words, the explanation is ``pointing'' to the correct region in the image. 

For \emph{Waterbirds-95\%} and \emph{Waterbirds-100\%}, we use segmentation masks of the birds for $\z$. On \emph{MSCOCO-ApparentGender}, we use the available person segmentation masks, choosing the mask with the largest bounding box if multiple people are present to be consistent with our task. Segmentation masks for red meat in Food-101 are not available. For generating model explanations, we use the black-box saliency method RISE~\cite{petsiuk2018rise}. \cref{fig:pg} presents our results. Our method matches the ABN baseline on \emph{MSCOCO-ApparentGender}. However, we outperform all baselines by 9.4\% on \emph{Waterbirds-95\%} and 12.8\% on \emph{Waterbirds-100\%}.

\subsection{Model Ablations}
\label{sec:ablations}

We explore several other design choices for the $VL$ model and attention method in Table~\ref{tab:design_choices}. We consider the Attention Branch Network (ABN) ~\cite{fukui2019attention} as the classification model, while supervising its feed-forward attention map (similar to ~\cite{mitsuhara2019embedding}). We also try supervising the GradCAM from the last convolutional layer of a ResNet50 classification model directly. For generating language specification, we experiment with the CLIP ViT-B/32 (CLIP ViT in the table). The method we denote as ``\method{}'' corresponds to the row with RRR as the classifier attention method supervised with CLIP ResNet50 GradCAM attention. 
For both ABN and GradCAM classifier attention methods, we compute $\mathcal{L}_{att}$ as an L1 loss in a similar style as in RRR --- penalizing $A^{f_{\theta}}$ where $A^{VL}$ is low, as opposed to matching $A^{f_{\theta}}$ directly to $A^{VL}$, finding that this gives slightly better performance. We chose RRR+CLIP R50 since it had the most consistent performance in minority class accuracy and fairness.

\section{Limitations and Broader Impacts}
\label{sec:conclusions}
In this work, we focus on a scenario where a dataset bias during training time is not present at test time. This is an important issue with serious implications for high-risk domains such as autonomous driving or medical imaging. Generally, as machine learning methods become widespread and impact people's lives, reliance on biases may be harmful to entire populations. Thus, we envision potential positive impact from our work towards mitigating this issue.

One of the datasets used in this work (\emph{MSCOCO-ApparentGender}) is derived from the image captioning MSCOCO-Bias and MSCOCO-Balanced splits introduced in~\cite{hendricks2018women}. Following \cite{hendricks2018women}, we consider three gender categories: male, female, and gender neutral (e.g., person) based on visual appearance. The gender labels were determined using a previously collected publicly released dataset in which annotators describe images~\cite{chen2015microsoft}. Importantly, people in the images are not asked to identify their gender. Thus, we emphasize that we are not classifying biological sex or gender identity, but rather outward gender appearance. In particular, we are interested in reducing gender entanglement with contextual features and ``equalizing'' the outcome across male and female categories.

We also would like to point out that in our experiments, we use the off-the-shelf large-scale vision-language model (CLIP~\cite{radford2021learning}) which may have encoded some internal biases, transferred from the data on which it was trained. Specifically, CLIP was trained on 400M image-caption pairs sourced from the Web, so we can not rule out the presence of biases or harmful (e.g. gender or racial) stereotypes in it. Practitioners who wish to use our approach should be mindful of such sources of bias.

\begin{figure}[b]
  \centering
  \includegraphics[width=0.85\linewidth]{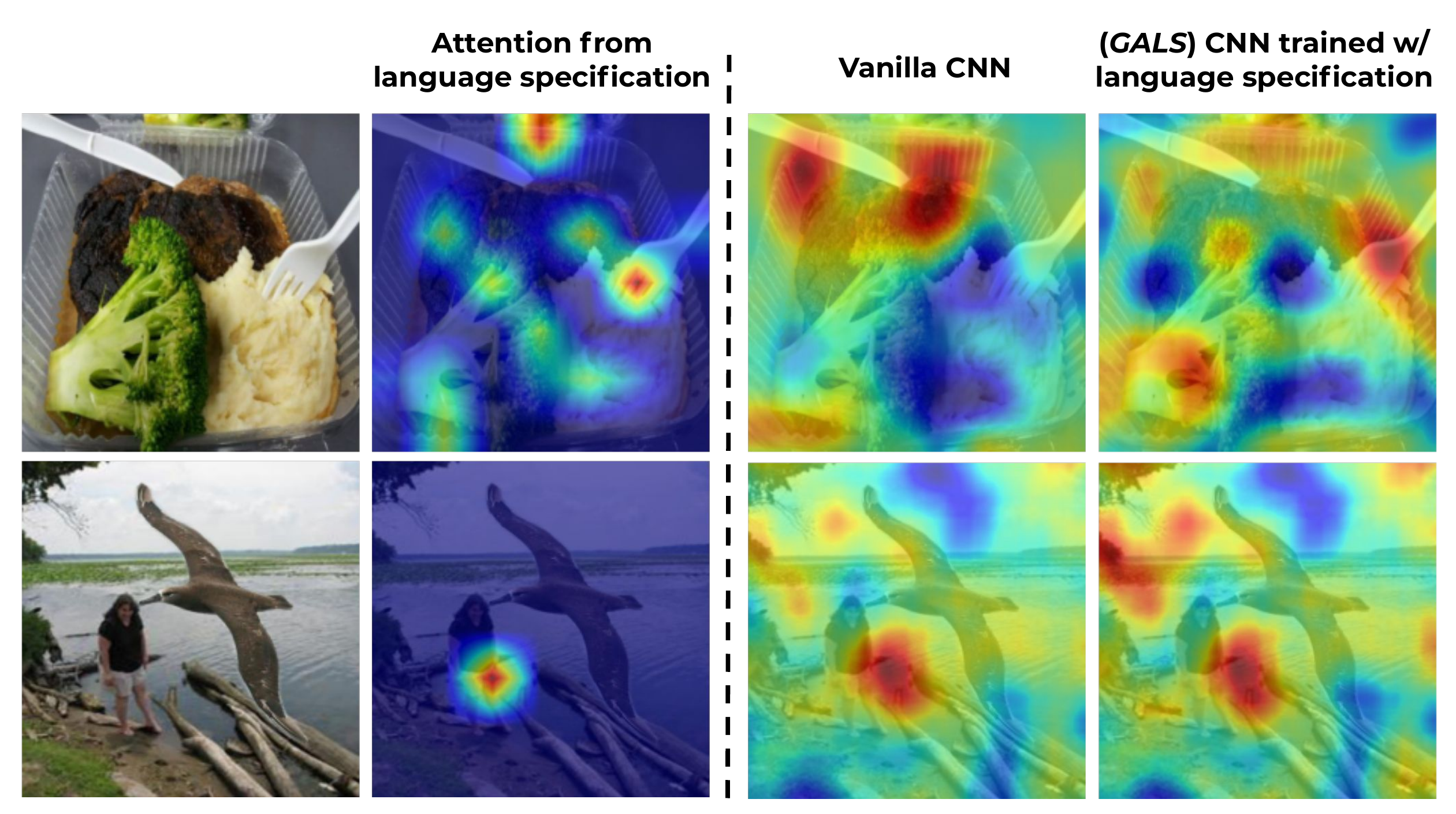}
  \caption{\textbf{Limitations.} Example of poor CLIP attention on  \emph{Red Meat} (top) and \emph{Waterbirds} (bottom) dataset. Since \method{} is supervised by the attention from language specification, our classifier's attention fails to ignore distractors when attention generated from language specification does not localize the task-relevant features.
  }
  \label{fig:limitations}
  \vspace{-3mm}
\end{figure}

As described in Sec.~\ref{sec:method}, our framework is limited to biases which can be pixel-wise separated from relevant features. As a counterexample, it would not apply to the task of classifying a person's age, with a confounding factor of race. Our framework also struggles when the vision and language model cannot ground the language specification (Fig.~\ref{fig:limitations}). In other scenarios, CLIP may struggle when the prompt is more compositional, such as ``the person in the blue shirt sitting next to the table''. 

\section{Acknowledgements}
We would like to thank Dr. Sayna Ebrahimi for helpful discussion. This work was supported in part by DoD, including DARPA's LwLL, and/or SemaFor programs, and Berkeley Artificial Intelligence Research (BAIR) industrial alliance programs. In addition to NSF CISE Expeditions Award CCF-1730628, this research is supported by gifts from Amazon Web Services, Ant Group, Ericsson, Facebook, Futurewei, Google, Intel, Microsoft, Scotiabank, and VMware.

{\small
\bibliographystyle{ieee_fullname}
\bibliography{main}
}

\clearpage
\appendix

We provide additional details on training and datasets. We also include several qualitative samples of attention from language specification, and compare Vanilla model attention to \method{} model attention.

\section{Training Details}
All runs were performed on 1-4 NVIDIA GeForce RTX 2080 GPUs. All models were optimized with stochastic gradient descent with a momentum of 0.9. For simplicity, we do not perform any data augmentations. For hyperparameter tuning, we split each dataset into training, validation, and testing, choosing the final hyperparameters based on which maximize validation accuracy. Our final hyperparameters are summarized in~\cref{tab:hp_table}. Language specifications used in the experiments are show in~\cref{tab:lang_specs}.

\textbf{\emph{Waterbirds-95\%}.} For the vanilla ResNet50 model, we perform a hyperparameter sweep with batch size 96, and run for 100 epochs. We sweep the backbone learning rate over 0.01, 0.005, 0.001, and 0.0001, and the linear classifier learning rate over 0.1, 0.01, 0.005, 0.001, and 0.0001. We chose a backbone learning rate of 0.01 and classifier learning rate of 0.001. For RRR, using the vanilla model learning rates, we first swept the attention loss weight ($\lambda$ in~\cref{eq:rrr}) over 1,000, 10,000, and 100,000, as well as the attention loss function over L1 and L2. From this, we chose a $\lambda$ of 10,000 and an L1 loss. Next, we ran the same learning rate sweep as for the vanilla model, and chose a backbone learning rate of 0.005 and classifier learning rate of 0.0001.

\textbf{\emph{Waterbirds-100\%}.} We use the same hyperparameters found for \emph{Waterbirds-95\%}.

\textbf{\emph{MSCOCO-ApparentGender}.} For the vanilla ResNet50 model, we run a hyperparameter sweep with a batch size of 96 for 100 epochs, testing backbone learning rates of  0.01, 0.005, and 0.001, and classifier learning rates of 0.1, 0.01, 0.005, and 0.001. We chose a backbone learning rate of 0.01 and classifier learning rate of 0.001. For attention weight $\lambda$, we test 1,000, 5,000, and 10,000, and choose 10,000 from validation.

\textbf{\emph{Red Meat}} For the vanilla ResNet50 model, we run a hyperparameter sweep with a batch size of 96 for 50 epochs, testing backbone learning rates of  0.1, 0.01, 0.001, 0.005, 0.001, 0.0001, 0.0005, and classifier learning rates of 0.1, 0.01, 0.001, 0.005, 0.001, 0.0001, and 0.0005. 
For attention weight $\lambda$, we test 100, 1,000, 10,000, and 100,000, and choose 10,000 from validation. 

\begin{table}[b]
  \centering
  \begin{tabular}{lrrr}
    \toprule
    \textbf{Split} & \textbf{Man} & \textbf{Woman} & \textbf{Person} \\
    \midrule
    Training & 10565 & 4802 & 2822 \\
    Validation & 500 & 500 & 0  \\
    Test & 500 & 500 & 0 \\
    \bottomrule
  \end{tabular}
  \caption{Dataset sizes on \emph{MSCOCO-ApparentGender}.}
 \label{tab:cocosizes}
\end{table}

\begin{table*}[h]
  \centering
  \begin{tabular}{llrrrrr}
    \toprule
     \textbf{Dataset} & \textbf{Method} & \textbf{Epochs} & \textbf{Batch Size} & \textbf{Base LR} & \textbf{Classifier LR} & \textbf{$\lambda$} \\
    \midrule
    \multirow{4}{*}{\emph{Waterbirds-95\%}} & Vanilla & 200 & 96 & 0.01 & 0.001 & - \\
    & ABN & 200 & 96 & 0.01 & 0.001 & - \\
    & UpWeight &  200 & 96 & 0.01 & 0.001 & - \\
    & \method{} & 200 & 96 & 0.005 & 0.0001 & 10,000\\
     \midrule
     \multirow{4}{*}{\emph{Waterbirds-100\%}} & Vanilla & 200 & 96 & 0.01 & 0.001 & - \\
     & ABN  & 200 & 96 & 0.01 & 0.001 & - \\
     & UpWeight & 200 & 96 & 0.01 & 0.001 & - \\
     & \method{} & 200 & 96 & 0.005 & 0.0001 & 10,000\\
     \midrule
      \multirow{2}{*}{\emph{Waterbirds-100\% Backgrounds}} & Vanilla & 200 & 96 & 0.01 & 0.001 & - \\
      & \method{} & 200 & 96 & 0.005 & 0.0001 & 1,000 \\
      \midrule
    \multirow{4}{*}{\emph{MSCOCO-ApparentGender}} & Vanilla  & 200 & 96 & 0.01 & 0.001 & - \\
    & ABN & 200 & 96 & 0.01 & 0.001 & - \\
    & UpWeight & 200 & 96 & 0.01 & 0.001 & - \\
    & \method{} & 200 & 96 & 0.01 & 0.001 & 10,000\\
     \midrule
    \multirow{3}{*}{\emph{Red Meat}} & Vanilla & 150 & 96 & 0.01 & 0.001 & - \\
    & ABN & 150 & 96 & 0.01 & 0.001 & - \\
    & \method{} & 150 & 96 & 0.001 & 0.001 & 10,000\\
    \bottomrule
  \end{tabular}
  \caption{Hyperparameter details. All models were optimized with SGD using a weight decay of $1e$-$5$. ``Base LR'' refers to the learning rate used for the pretrained ResNet50 backbone, and ``Classifier LR'' refers to the learning rate used for the linear classifier. $\lambda$ is the attention loss weight from in~\cref{eq:rrr}.}
  \label{tab:hp_table}
\end{table*}

\begin{table*}[h]
  \centering
  \begin{tabular}{ll}
    \toprule
    \textbf{Dataset} & \textbf{Language specifications}\\
    \midrule
\emph{Waterbirds-95\%} & ``\{a photo/an image\} of a bird''\\
\midrule
\emph{Waterbirds-100\%} &  ``\{a photo/an image\} of a bird''\\
\midrule
\emph{Waterbirds-100\% Backgrounds} & \makecell[l]{``\{a photo/an image\} of a nature scene'',  ``\{a photo/an image\} of an outdoor scene'', \\ ``\{a photo/an image\} of a landscape''}\\
\midrule
\emph{MSCOCO-ApparentGender} & ``\{a photo/an image\} of a person''\\
\midrule
\emph{Red Meat} & ``\{a photo/an image\} of meat''\\
    \bottomrule
  \end{tabular}
  \caption{Language specifications used for \method{} in experiments. ``\{a photo/an image\} of X'' indicates that two prompts were used: ``a photo of X'' and ``an image of X''.}
  \label{tab:lang_specs}
\end{table*}

\begin{table*}[h]
  \centering
  \begin{tabular}{lrrrr}
    \toprule
    \textbf{Split} & \textbf{Landbirds, land} & \textbf{Landbirds, water} & \textbf{Waterbirds, land} & \textbf{Waterbirds, water} \\
    \midrule
    Training & 3498 & 184 & 56 & 1057 \\
    Validation & 467 & 466 & 133 & 133 \\
    Test & 2255 & 2255 & 642 & 642\\
    \bottomrule
  \end{tabular}
  \caption{Dataset sizes on \emph{Waterbirds-95\%}. The two classes are ``Landbird'' and ``Waterbird.'' Furthermore, each image can display either a land background or a water background.}
  \label{tab:waterbirdssizes95}
\end{table*}

\begin{table*}[h]
  \centering
  \begin{tabular}{lrrrr}
    \toprule
    \textbf{Split} & \textbf{Landbirds, land} & \textbf{Landbirds, water} & \textbf{Waterbirds, land} & \textbf{Waterbirds, water} \\
    \midrule
    Training & 3694 & 0 & 0 & 1101 \\
    Validation & 467 & 466 & 133 & 133 \\
    Test & 2255 & 2255 & 642 & 642\\
    \bottomrule
  \end{tabular}
  \caption{Dataset sizes on \emph{Waterbirds-100\%}. The validation and test splits have the same distribution as validation and test in Table \ref{tab:waterbirdssizes95} for \emph{Waterbirds-95\%}.}
  \label{tab:waterbirdssizes100}
\end{table*}

\begin{table*}[h]
  \centering
  \begin{tabular}{lrrrrr}
    \toprule
    \textbf{Split} & \textbf{Filet Mignon} & \textbf{Filet Mignon} & \textbf{Pork Chop} & \textbf{Prime Rib} & \textbf{Steak}\\
    \midrule
    Training & 500 & 500 & 500 & 500 & 500 \\
    Validation & 250 & 250 & 250 & 250 & 250 \\
    Test & 250 & 250 & 250 & 250 & 250 \\
    \bottomrule
  \end{tabular}
  \caption{Dataset sizes on \emph{Food-101}.}
  \label{tab:food_sizes}
\end{table*}

\section{Dataset Details}
\textbf{Waterbirds variants.} For \emph{Waterbirds-95\%}, we use the same dataset as provided by the authors of ~\cite{sagawa2019distributionally}. For \emph{Waterbirds-100\%}, we follow the code provided by those authors for generating a new synthetic dataset. Land backgrounds are randomly chosen from the ``bamboo forest'' and ''broadleaf forest'' categories in the Places dataset, and water background are from the ``ocean'' and ``natural lake'' categories. These categories were determined in ~\cite{sagawa2019distributionally}. Both dataset variants have $4,795$ training images, $1,119$ validation images, and $5,794$ test images. Tables \ref{tab:waterbirdssizes95} and \ref{tab:waterbirdssizes100} show the number of samples per class, broken down further by the type of background.

\textbf{MSCOCO-ApparentGender}. For the training set, we begin by using the $22,966$ MSCOCO image ids defined in the Bias split in ~\cite{zhao2017men}. We next filter and label these images using a list of ``male'' words (such as ``father'', ``man'', or ``groom''), a list of ``female'' words (such as ``daughter'', ``lady'', or ``she''), and a list of ``person'' words which do not have a defined gender (such as ``child'', ``surfer'' or ``employee'') provided by ~\cite{hendricks2018women}. From these provided lists, we filter out plural words. Next, we filter out images where the annotators do not agree on the gender (at least one caption mentions a male word and at least one caption mentions a female word). We label an image as ``Man'' if the majority of annotators (3 out of the 5 available captions per image) mention a male word, and ``Woman'' if the majority mention a female word. We label an image as ``Person'' if it has not been labeled as ``Man'' or ``Woman'' and if the majority of annotators have mentioned a ``person'' word. We use the same validation and test images for ``Man'' and ``Woman'' as in the ``Balanced'' split defined in ~\cite{hendricks2018women}. Although these were not labeled in the same manner as our training set, we keep the splits the same to have consistent evaluation with prior work. The number of samples per class is summarized in Table \ref{tab:cocosizes}.

\textbf{Food-101.} We start by selecting the 5 red meat classes from the Food-101 dataset~\cite{food_101} and split the 750 training samples into 500 training samples and 250 validation samples, keeping the 250 sample test set the same. The number of samples per class is summarized in Table \ref{tab:food_sizes}.

\subsection{Attention Samples}

\begin{figure}
  \centering
  \includegraphics[width=\linewidth]{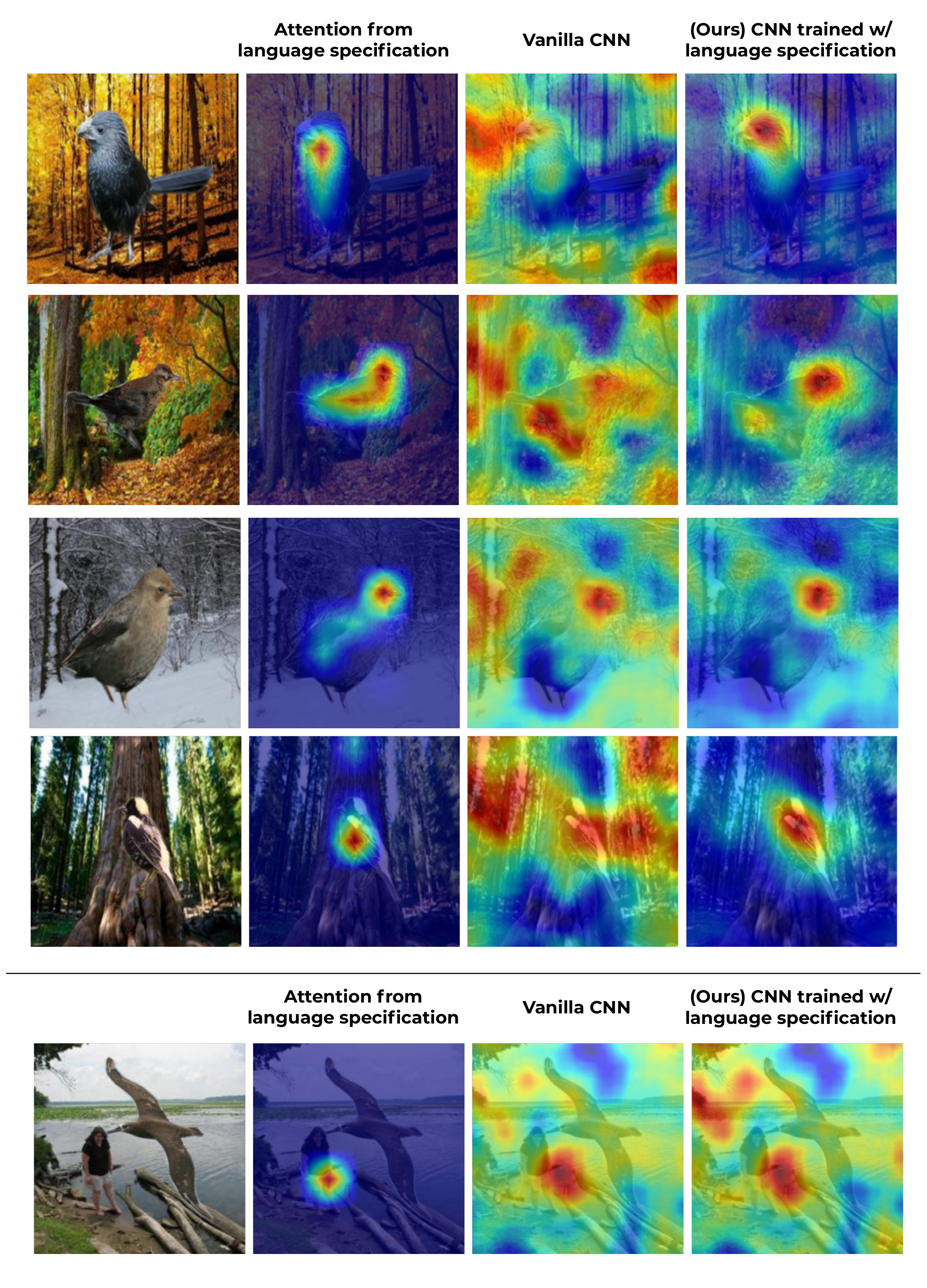}
  \caption{Sample attention visualizations from the \emph{Waterbirds-100\%} training set. Our model places considerably less attention on the background features than did the Vanilla baseline. However, it can have difficulty localizing the bird in cases where the language-guided attention also has difficulty in grounding, as shown in the bottom row.}
  \label{fig:wbird_samples}
\end{figure}

\begin{figure}
  \centering
  \includegraphics[width=\linewidth]{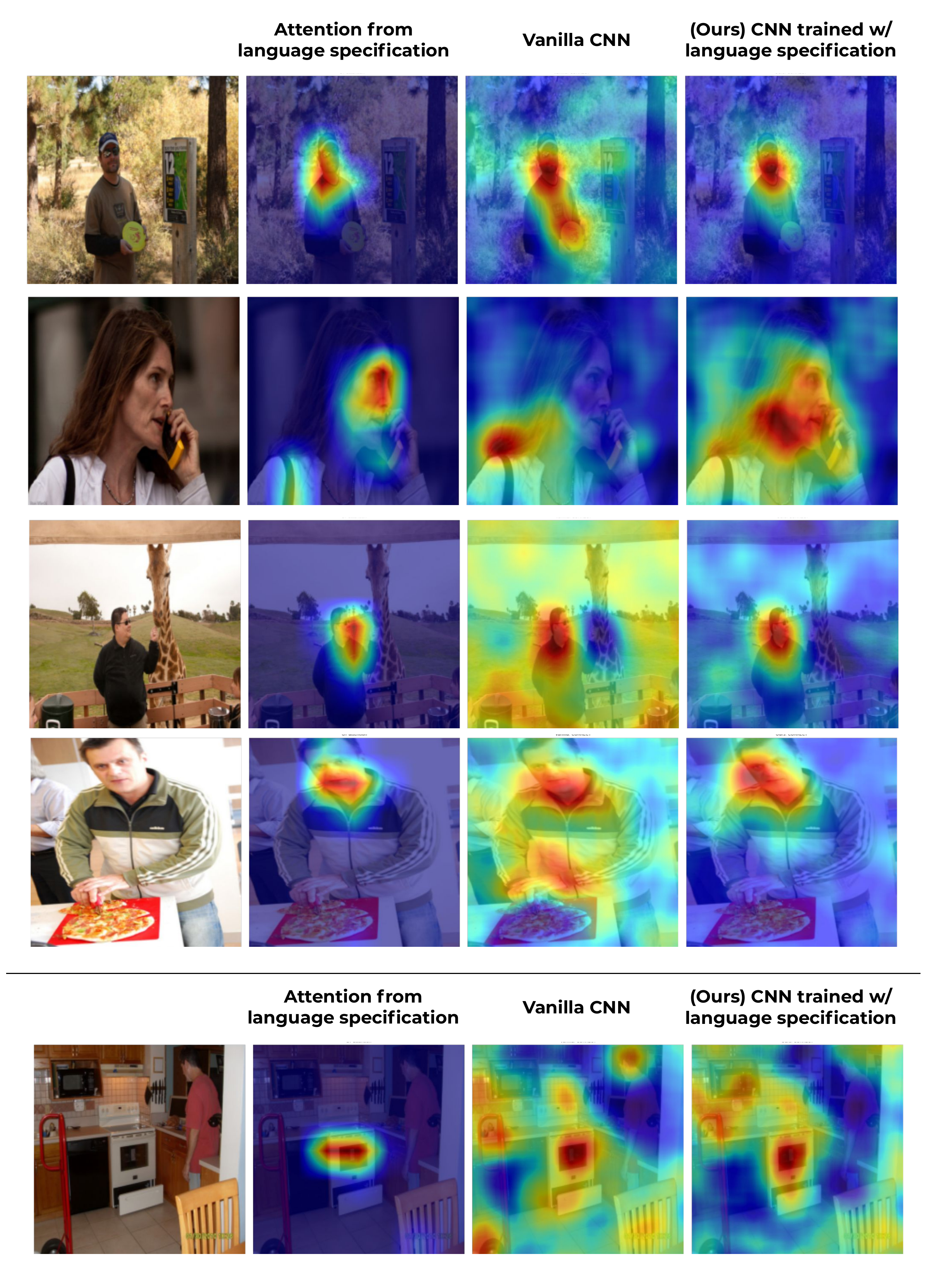}
    \caption{Sample attention visualizations from the \emph{MSCOCO-ApparentGender} training set. Like the attention from language specification, our model is proficient at identifying faces, and placing less attention on potentially biased context. However, the sample shown in the bottom row is an example where the language-guided attention does not localize the person correctly, and our model attends to similar features as the vanilla model.}
  \label{fig:coco_samples}
\end{figure}

\begin{figure}
  \centering
  \includegraphics[width=\linewidth]{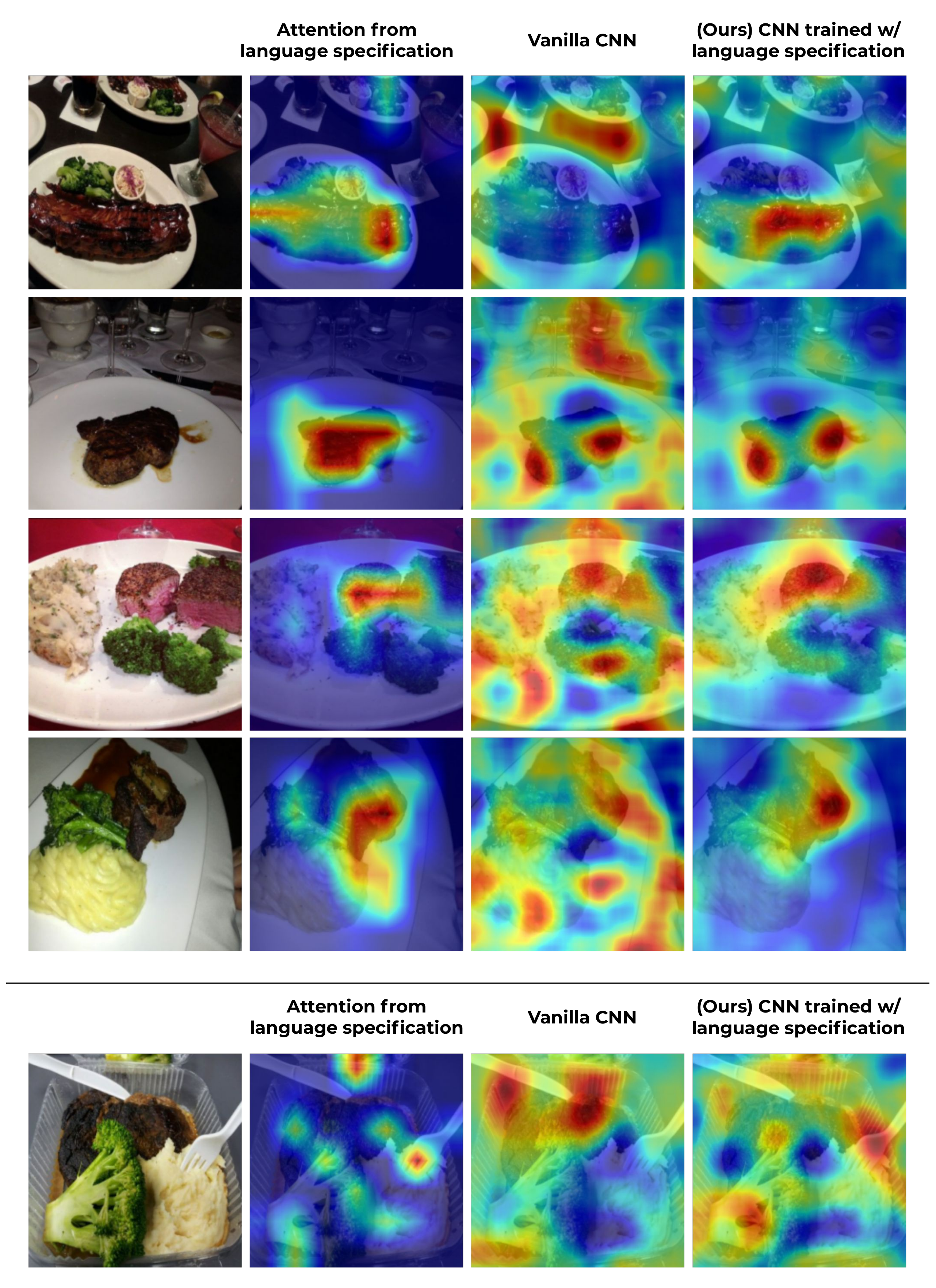}
    \caption{Sample attention visualizations from the \emph{Red Meat} training set. The images tend to show cluttered plates of food, where the meat is often not centered. \method{} can learn to attend to the meat itself when guidance from the language specification is also able to localize the meat.}
  \label{fig:food_samples}
\end{figure}

In Figures \ref{fig:wbird_samples}, \ref{fig:coco_samples}, and \ref{fig:food_samples}, we show several qualitative examples of spatial attention. Specifically, for sample images from the \emph{Waterbirds-100\%}, \emph{MSCOCO-ApparentGender}, and \emph{Food-101} training sets, we show the CLIP ResNet50 GradCAM $A^{VL}$ guidance, as well as the RISE attention for the vanilla model and ours. We show that in many cases, our model has learned to attend to similar image features as the language-guided attention. However, when the image is difficult for the language-guided attention to ground the object of interest, then our model can have more difficulty in localization as well.

 \begin{table*}[h]
  \centering
  \label{tab:waterbirds}
  \begin{tabular}{lrrrrr}
    \toprule
      & \multicolumn{2}{c}{\textbf{\emph{Waterbirds 95\%}}}  & & \multicolumn{2}{c}{\textbf{\emph{Waterbirds 100\%}}}                 \\
  \cmidrule(r){2-3}\cmidrule(r){5-6}
      \textbf{Method}  & \textbf{Per Group} & \textbf{Worst Group} &  & \textbf{Per Group} & \textbf{Worst Group}\\
    \midrule
        CLIP Zero-shot &	73.18 &	43.46 & & 73.18 &	43.46\\
        CLIP Finetune, LogisticReg.  &	80.58 &	56.85 & & 	68.36 &	32.15 \\
        \midrule
      Vanilla  & 	86.93 $\pm$ 0.46 &	73.07 $\pm$ 2.24  & &  69.83 $\pm$ 2.04 & 34.31$\pm$ 7.31\\
      UpWeight Class &	86.74 $\pm$ 0.54  &	73.66 $\pm$ 2.00  & & 	70.50 $\pm$ 2.00 &	34.82 $\pm$ 6.65 \\
      ABN & 86.01 $\pm$ 0.70 & 65.03 $\pm$ 2.77 & & 72.20 $\pm$ 3.02 & 41.56 $\pm$ 6.76 \\
      \method{}  & 	\textbf{89.05 $\pm$ 0.47} &	\textbf{76.54 $\pm$ 2.40}  & & \textbf{79.72 $\pm$ 1.60}	& \textbf{56.71 $\pm$ 3.92} \\
    \bottomrule
  \end{tabular}
  \caption{Test accuracy of approaches on the \emph{Waterbirds-95\%} and \emph{Waterbirds-100\%} datasets. The percentage indicates the proportion of training samples in each class which have a spurious correlation between the class label and the background. }
\end{table*}

 \section{Results Tables}
 Here we provide the tabular results for each of the figures in ~\cref{sec:experiments} of the main paper. 

\begin{table}[h]
  \centering
  \begin{tabular}{lrrr}
    \toprule
                        &  \multicolumn{3}{c}{\textbf{Pointing Game Accuracy}}\\
                        \cmidrule(r){2-4}
     \textbf{Method}    & \textbf{Man} &  \textbf{Woman} &  \textbf{Average} \\
    \midrule
     Vanilla & 51.20 &  64.40  & 57.80 \\
      ABN & \underline{55.80} & \textbf{69.60} & \textbf{62.70}  \\
      UpWeight & 42.60 & 57.00 & 49.80 \\
      \textbf{\method{}} & \textbf{56.20} & \underline{67.00} & \underline{62.60} \\
    \bottomrule
  \end{tabular}
  \caption{Pointing game accuracy on \emph{MSCOCO-ApparentGender}.}
  \label{tab:pg_gender}
\end{table}

\begin{table}[t]
  \centering
  \begin{tabular}{lrrr}
    \toprule
        & \multicolumn{2}{c}{\textbf{Pointing Game Accuracy}}\\
        \cmidrule(r){2-3}
     \textbf{Method} & \textbf{\emph{Waterbirds-95\%}} & \textbf{\emph{Waterbirds-100\%}} \\
    \midrule
      Vanilla & \underline{59.98} & \underline{46.48}\\
      ABN & 51.73 & 25.96 \\
      UpWeight & 59.42 & 26.34 \\
      \textbf{\method{}} & \textbf{69.38} & \textbf{59.27}\\
    \bottomrule
  \end{tabular}
  \caption{Pointing game accuracy on \emph{Waterbirds} datasets.}
  \label{tab:pg_waterbirds}
\end{table}

\end{document}